\documentclass{article}

\PassOptionsToPackage{table}{xcolor}
\usepackage[main, final]{neurips_2026}

\makeatletter
\renewcommand{\@notice}{}
\makeatother

\usepackage[utf8]{inputenc}
\usepackage[T1]{fontenc}
\usepackage{hyperref}
\usepackage{url}
\usepackage{booktabs}
\usepackage{amsfonts}
\usepackage{nicefrac}
\usepackage{microtype}
\usepackage{xcolor}

\usepackage{graphicx}
\usepackage{amsmath}
\usepackage{amssymb}
\usepackage{algorithm}
\usepackage{algorithmic}
\usepackage{multirow}
\usepackage{makecell}

\title{Agentic Visual Reasoning in Whole-Slide Pathology Images via Active Perception}
\author{%
  \textbf{Jingyun Chen}\textsuperscript{1,*} \quad
  \textbf{Fengchun Liu}\textsuperscript{1,*} \quad
  \textbf{Linghan Cai}\textsuperscript{1} \quad
  \textbf{Songhan Jiang}\textsuperscript{1} \\
  \textbf{Shenjin Huang}\textsuperscript{1} \quad
  \textbf{Hongpeng Wang}\textsuperscript{1,$\dagger$} \quad
  \textbf{Lequan Yu}\textsuperscript{2,$\dagger$} \quad
  \textbf{Yongbing Zhang}\textsuperscript{1,$\dagger$}
  \\
  \normalfont\textsuperscript{1}College of Computer Science and Technology, Harbin Institute of Technology, Shenzhen \\
  \textsuperscript{2}School of Computing and Data Science, The University of Hong Kong \\
  \texttt{\{jychen, cailh\}@stu.hit.edu.cn, ybzhang08@hit.edu.cn}
}

\begin{document}
\maketitle

\begingroup
\renewcommand{\thefootnote}{\fnsymbol{footnote}}
\footnotetext[1]{Equal contribution.} \footnotetext[2]{Corresponding authors.}
\endgroup

\begin{abstract}
Whole-slide visual reasoning requires identifying sparse diagnostic evidence in gigapixel pathology slides and integrating observations across spatial scales. Existing WSI methods either compress densely sampled patches into global representations or rely on pretrained vision-language models and heuristic planning to select regions. The former weakens the connection between predictions and the morphological evidence, whereas the latter lacks a pathology-trained policy for deciding where and over what spatial extent to observe. We present AdaptivePath, an active-perception framework that formulates WSI evidence acquisition as a sequential decision-making process. Since question-specific evidence trajectories are costly to annotate, the Navigator learns question-agnostic, abnormality-driven navigation from pathologist-reviewed labels to select the location and spatial extent of each observation. We train this policy by alternating representation learning with proximal policy optimization and subsequently fine-tune it using geometric and appearance consistency objectives to stabilize focus trajectories under visual variations. During inference, the Navigator hierarchically acquires sparse observations from low to high magnification under a limited ROI budget. A Morphology Interpreter converts these observations into question-conditioned morphological evidence, allowing the Deliberator to evaluate newly acquired evidence and revise intermediate answers across magnifications. The Arbiter then integrates the resulting deliberation history to produce the final answer. AdaptivePath achieves state-of-the-art zero-shot performance on WSI-level and region-level pathology VQA benchmarks and an overall accuracy of 80.14\% for cancer subtype classification across six TCGA cohorts. In a blinded diagnostic-utility study, pathologists using AdaptivePath-selected observation sequences achieve 82.9\% accuracy. These results demonstrate that learned active perception provides an effective and traceable basis for agentic visual reasoning over gigapixel pathology slides. The code is available at \url{https://github.com/G14nTDo4/AdaptivePath}.
\end{abstract}

\section{Introduction}
Whole-slide visual reasoning is an agentic perception problem. Unlike conventional visual question answering across bounded images~\cite{vqa,vilt}, reasoning over a whole-slide image (WSI) requires operating on billions of pixels that depict heterogeneous tissue structures and sparse diagnostic findings~\cite{campanella,clam,rlogist}. An agent cannot inspect such an image exhaustively at its native resolution. Instead, it must select the location and spatial extent of each observation, retain acquired evidence, and integrate observations across magnifications to answer a question~\cite{slideseek,pathologycot}. Active visual evidence acquisition is therefore fundamental to agentic reasoning over gigapixel pathology slides.

Recent systems such as PathAgent, PathNavigate, and GIANT acquire visual evidence using pretrained retrieval models, heuristic search, or language-model planning~\cite{pathagent,pathnavigate,giant}, while CPathAgent learns multi-scale navigation from model-generated trajectories~\cite{sun2026cpathagent}. Their navigation depends on pretrained models or generated plans, and their reasoning lacks explicit option-wise evidence auditing. Direct VQA rewards do not indicate which selected regions contain the evidence, whereas question-specific evidence trajectories are costly to annotate~\cite{e2mil,geomil,pathagentbench}. Effective WSI reasoning therefore requires a navigation policy grounded in pathology that can be reused across questions, together with explicit evaluation of the acquired evidence for each candidate answer.

To meet these requirements, we present AdaptivePath, which formulates WSI visual reasoning as a scale-progressive active perception process. At each magnification, the Navigator scores candidate regions, predicts the locations and spatial extents of observations for the Interpreter, and retains top-ranked candidates for the next search space. The Interpreter generates morphological descriptions conditioned on the question, and the Deliberator then audits how directly observed morphology supports or conflicts with each option and assesses whether the evidence is sufficient. Rather than selecting an answer by voting over magnification-specific predictions, the Arbiter evaluates the accumulated region-level morphological descriptions and evidence audits to produce the final answer. We optimize the Navigator using pathologist-reviewed abnormality labels. These components establish an explicit and traceable path from WSI exploration to final adjudication. Our main contributions are as follows:

\begin{itemize}
    \item We introduce AdaptivePath, an active-perception system for agentic WSI visual reasoning. The system integrates progressive navigation across magnifications, morphology interpretation conditioned on the question, evidence deliberation at each magnification, and final adjudication across magnifications into an explicit path from visual exploration to the final decision.

    \item We construct an abnormal-region recognition dataset with pathologist-reviewed labels and use it to train a Navigator that localizes abnormalities, estimates their spatial extents, and provides observations across multiple magnifications to the downstream reasoning modules.

    \item We comprehensively evaluate AdaptivePath on pathology VQA, cancer subtype classification, a pathologist-centered diagnostic utility test, and navigation trajectory alignment, assessing its question-answering performance, generalization, diagnostic value, and navigation behavior.
\end{itemize}

\section{Related Work}
\subsection{Pathology and WSI Vision-Language Models}
General-purpose vision-language models~\cite{clip,llava,qwenvl,blip2} have motivated pathology-specific extensions. PLIP~\cite{plip} and CONCH~\cite{conch} learn aligned representations of pathology images and text, while Quilt-LLaVA~\cite{quiltllava}, PathChat~\cite{pathchat}, PathAsst~\cite{pathasst} and PathGen~\cite{pathgen} further connect pathology visual encoders with large language models through instruction tuning. Patho-R1~\cite{pathor1} and ScaleReasoner-R1~\cite{scalereasoner-r1} enhance visual reasoning through chain-of-thought supervision and reinforcement learning. However, these models primarily operate on region-level pathology images. WSI-VQA~\cite{wsi-vqa}, SlideChat~\cite{slidechat} and WSI-LLaVA~\cite{wsi-llava} extend language interaction to gigapixel WSIs by connecting pre-extracted patch features to language models, but their visual reasoning relies on compressed representations rather than explicit sequences of region-level observations.

\begin{figure}[htbp]
    \centering
        \includegraphics[width=1\textwidth]{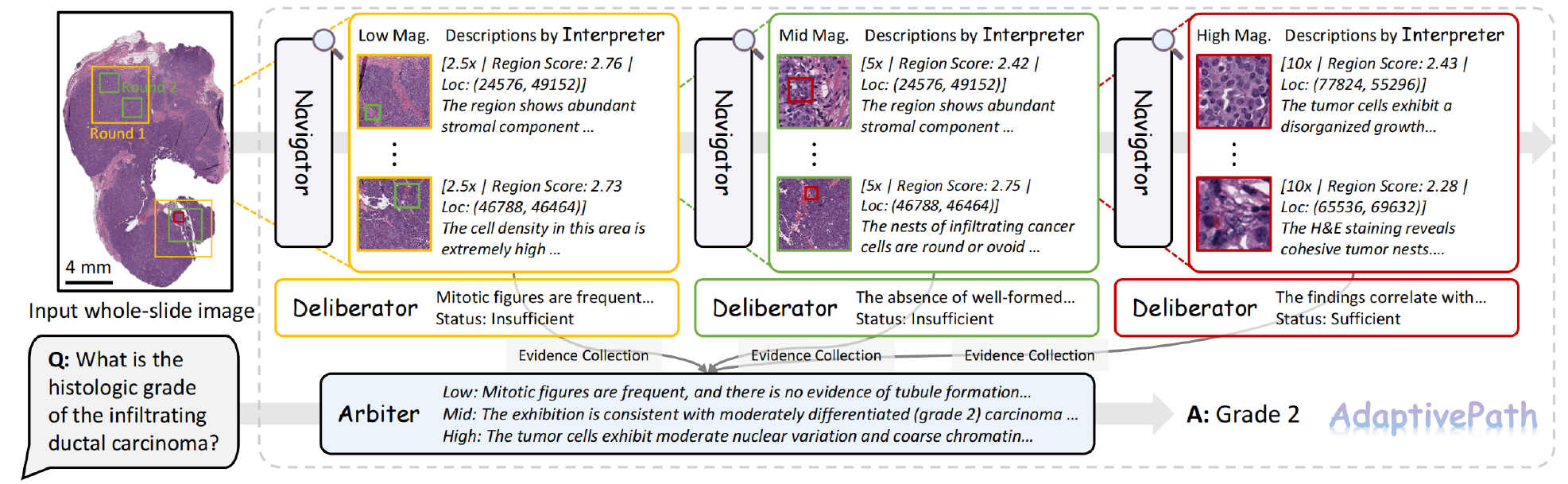}
    \caption{Overview of AdaptivePath. The Navigator is trained on abnormal-region recognition data to sequentially predict focus locations and scales and estimate abnormality. During inference, AdaptivePath hierarchically acquires and integrates multi-scale morphological evidence to produce the final answer.}
    \label{fig:overview}
\end{figure}

\subsection{Agentic WSI Visual Reasoning}
To move beyond reasoning over static patch features, recent agentic WSI systems explicitly acquire visual evidence through iterative region selection and multi-step reasoning. PathAgent~\cite{pathagent} uses a Navigator guided by similarity scores from a pretrained vision-language model to retrieve candidate regions and employs multi-step reasoning driven by a language model to adjust the magnification and guide subsequent region retrieval. PathNavigate~\cite{pathnavigate} constructs a slide-specific surprise prior from frozen pathology features and uses alignment scores from PLIP to rerank regions according to their relevance to the question. GIANT~\cite{giant} instead employs a general-purpose multimodal model to iteratively plan and navigate using downsampled slide observations. However, these systems have limited supervision for abnormal-region localization and lack explicit option-wise auditing of directly observed evidence across magnifications.

\subsection{Active Perception for Visual Reasoning}
Active perception ranges from coarse-to-fine fixation selection in AdaptiveNN~\cite{adaptivenn} to agentic reasoning through environmental navigation or task-relevant region selection~\cite{wang2023active,zhou2024navgpt}. Applying active perception to WSIs is challenging because diagnostically relevant morphological features varie in location and spatial extent and may become apparent at different magnifications across a gigapixel image. CPathAgent~\cite{sun2026cpathagent} learns multi-scale navigation through a three-stage training pipeline using model-generated trajectories. However, its navigation supervision is derived from generated trajectories rather than dedicated pathologist-reviewed region labels. AdaptivePath addresses this limitation by using pathologist-reviewed abnormality labels to optimize a sequential policy that determines the location and spatial extent of each observation, thereby yielding pathology-grounded visual evidence for downstream reasoning.

\section{Methodology}
\label{sec:method}

\subsection{Overview}
\label{sec:method_overview}
As illustrated in Figure~\ref{fig:overview}, AdaptivePath is an agentic framework that integrates active perception, morphology interpretation, evidence deliberation, and final adjudication for whole-slide visual reasoning. The Navigator performs hierarchical active perception across magnifications by selecting the location and spatial extent of each observation, while the Interpreter converts these observations into morphological descriptions conditioned on the question. The Deliberator organizes the resulting evidence for candidate options, and the Arbiter directly adjudicates the evidence across magnifications to generate the final answer.

\begin{figure}[htbp]
    \centering
    \includegraphics[width=1\textwidth]{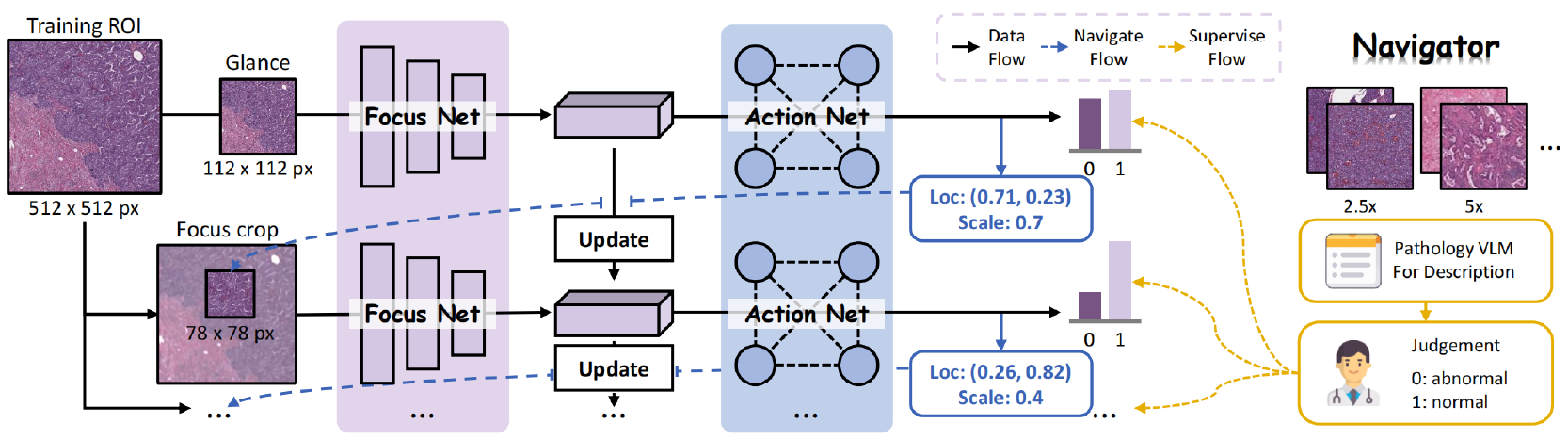}
    \caption{Overview of the active-perception Navigator. FocusNet encodes the global glance and sequential focus regions to update the visual state, while ActionNet predicts region-level abnormality logits and the location and spatial extent of the next observation. The Navigator is trained with pathologist-reviewed labels for abnormality-aware navigation.}
    \label{fig:navigator}
\end{figure}

\subsection{Active Perception Navigator}
\label{sec:navigator_training}

To enable abnormality-aware and scale-adaptive navigation, we train the Navigator using a multi-magnification abnormal-region recognition dataset as illustrated in Figure~\ref{fig:navigator}. Its construction details are provided in Section~\ref{sec:dataset-construction}. The Navigator is trained in two successive phases. Phase I alternates representation learning and policy optimization, while Phase II refines the resulting policy with geometric and appearance consistency.

\subsubsection{Sequential Focus Observation}
The Navigator $F_N$ consists of a FocusNet $\pi_F$ and an ActionNet $\pi_\theta$. Given a pathology region $I\in\mathbb{R}^{3\times P\times P}$, FocusNet encodes its globally resized glance $I_0$ to initialize the visual state $h_0$. At step $t$, ActionNet predicts the current abnormality logits $z_{t-1}$ and the next focus action:
\begin{equation}
    z_{t-1},a_t=\pi_\theta(h_{t-1}),
    \label{eq:action}
\end{equation}
where $h_{t-1}$ summarizes the observations acquired before step $t$. The action $a_t=(u_t,v_t,s_t)$ specifies the location and relative spatial extent of the next focus region. After the corresponding image region $I_t$ is acquired, FocusNet updates the visual state as
\begin{equation}
    h_t=\pi_F(h_{t-1},I_t).
    \label{eq:focus}
\end{equation}
The resulting state $h_t$ guides the next action. After the final observation,
the terminal logits are $z_T=\pi_\theta^{\mathrm{cls}}(h_T)$.

\subsubsection{Phase I: Representation and Perception Learning}
The Navigator needs to learn both discriminative representations for abnormality recognition and an observation policy that actively acquires informative observations. Following AdaptiveNN~\cite{adaptivenn}, we alternate between representation learning and policy optimization. The glance and all focus outputs receive cross-entropy supervision from the ground-truth navigation label $y^{\mathrm{nav}}$. The final focus prediction provides a stop-gradient teacher distribution for the glance and preceding focus predictions. Let $p_t=\operatorname{softmax}(z_t)$ and $\bar{p}_T=\operatorname{stopgrad}(p_T)$. The representation objective is
\begin{equation}
\mathcal{L}_{\mathrm{repr}} =
\sum_{t=0}^{T}\mathcal{L}_{\mathrm{CE}}(z_t,y^{\mathrm{nav}})+\alpha\sum_{t=0}^{T-1}
{D}_\mathrm{KL}(\bar{p}_T\Vert p_t).
\label{eq:repr_loss}
\end{equation}
The stopgrad operation prevents the teacher distribution from being updated by the distillation term.

To encourage informative focus observations, the immediate reward is defined by the reduction in classification loss:
\begin{equation}
    g_t=\mathcal{L}_{\mathrm{CE}}(z_{t-1},y^{\mathrm{nav}})
        -\mathcal{L}_{\mathrm{CE}}(z_{t},y^{\mathrm{nav}}).
    \label{eq:reward}
\end{equation}
Thus, an observation receives a positive reward when it improves abnormality recognition. 

During policy optimization, the visual FocusNet backbone $\pi_F$ is frozen, and the ActionNet policy $\pi_\theta$ is trained using generalized advantage estimation and clipped PPO~\cite{ppo}. The overall policy objective is
\begin{equation}
\mathcal{L}_{\mathrm{PPO}} =
\mathcal{L}_{\mathrm{clip}}
+\mathcal{L}_{V}
+\beta_{\mathrm{stab}}\mathcal{L}_{\mathrm{stab}}
-\beta_{\mathrm{ent}}\operatorname{Ent}(\pi_\theta),
\label{eq:ppo_loss}
\end{equation}
where $\mathcal{L}_{\mathrm{clip}}$ is the standard clipped surrogate loss, $\mathcal{L}_{V}$ is the value regression loss, $\mathcal{L}_{\mathrm{stab}}$ stabilizes representations across PPO updates, and the entropy term $\operatorname{Ent}(\pi_\theta)$ encourages exploration. Alternating representation and policy updates enable the Navigator to recognize potentially abnormal morphology and adaptively select the location and spatial extent of each observation.

\subsubsection{Phase II: Consistency-Regularized Policy Fine-Tuning}
Starting from the Phase I checkpoint, Phase II further stabilizes navigation by enforcing consistent localization under pathology-preserving geometric and appearance variations. For each region, we obtain deterministic focus trajectories from the original, geometrically transformed, and appearance-augmented views, with geometric predictions mapped back to the original coordinate system.

Each predicted focus box is converted into a normalized Gaussian evidence heatmap. Let $H_t$ denote the heatmap at step $t$ and $H$ the aggregate heatmap over all steps. The consistency objective is
\begin{equation}
\mathcal{L}_{\mathrm{cons}}=
D_{\mathrm{JS}}(H\Vert H^{\mathrm{geo}})
+\frac{1}{T}\sum_{t=1}^{T}
D_{\mathrm{JS}}(H_t\Vert H_t^{\mathrm{geo}})+D_{\mathrm{JS}}(H\Vert H^{\mathrm{app}}),
\label{eq:consistency}
\end{equation}
where $^\mathrm{geo}$ and $^\mathrm{app}$ denote geometric and appearance augmentations, respectively. The geometric terms preserve both step-wise trajectory correspondence and aggregate localization, while the appearance term stabilizes the overall focus distribution under visual variations. Phase II therefore fine-tunes the policy using $\mathcal{L}_{\mathrm{PPO}}+\mathcal{L}_{\mathrm{cons}}$.

\subsection{Agentic Whole-Slide VQA}
\subsubsection{Hierarchical Active Navigation}
\label{sec:hierarchical_navigation}

During inference, AdaptivePath applies the learned active perception policy hierarchically across the ordered magnifications $\mathcal{M}$. Let $\mathcal{R}_m$ denote the candidate region set at magnification $m$. The initial candidates cover the entire WSI, while subsequent candidates are generated within the regions retained at the preceding magnification. For each $R^m\in\mathcal{R}_m$, the Navigator acquires $T$ focus observations and computes an abnormality score from the final logits:
\begin{equation}
    \operatorname{AS}(R^m)=z_{T}^{m}[\mathrm{abnormal}]
             -z_{T}^{m}[\mathrm{normal}].
    \label{eq:abnormal_score}
\end{equation}
At each magnification, all candidate regions are ranked by their abnormality scores. We retain
\begin{equation}
    \mathcal{S}_m
    =
    \underset{R^m \in \mathcal{R}_m}{\operatorname{Top}_K}
    \,\operatorname{AS}(R^m),
\end{equation}
where $K$ is the navigation budget. The retained regions $\mathcal{S}_m$ define the next-magnification search space, while their predicted focus regions provide inputs to the Interpreter.

\subsubsection{Morphology Interpretation and Evidence Adjudication}
Given question $q$, let $\{f_{i,j}^m\}_{j=1}^{T}$ denote the $T$ focus observations acquired from selected region $R_i^m\in\mathcal{S}_m$. The Interpreter converts these observations into question-relevant descriptions $\{\operatorname{Des}(f_{i,j}^m)\}_{j=1}^{T}$ of directly visible morphology. The Deliberator aggregates these descriptions into a region-level summary of directly observed morphology:
\begin{equation}
    \gamma_i^m=
    F_D^{\mathrm{aggregate}}
    \left(
    q,
    \left\{
    \operatorname{Des}(f_{i,j}^m)
    \right\}_{j=1}^{T},
    m
    \right).
    \label{eq:region_evidence_aggregation}
\end{equation}
Pairing each summary with the corresponding region abnormality score gives the region morphology set
\begin{equation}
    \mathcal{G}_m=
    \left\{
    \left(
    \operatorname{AS}(R_i^m),
    \gamma_i^m
    \right)
    \,\middle|\, R_i^m\in\mathcal{S}_m
    \right\}.
    \label{eq:region_morphology_set}
\end{equation}

Based on the region morphology set, the Deliberator first produces an independent answer and rationale:
\begin{equation}
    O_m=F_D^{\mathrm{answer}}(q,\mathcal{G}_m,m).
    \label{eq:independent_level_answer}
\end{equation}

The Deliberator then reviews the evidence for every answer option and assesses overall evidence sufficiency:
\begin{equation}
    \left(\mathcal{A}_m,U_m\right)
    =
    F_D^{\mathrm{review}}(q,\mathcal{G}_m,O_m,m),
    \label{eq:evidence_sufficiency_review}
\end{equation}
where $\mathcal{A}_m$ records supporting findings and explicit conflicts for every option based solely on directly observed morphology, while absent or unobserved features are treated as neutral. $U_m$ records evidence sufficiency and its rationale.

The Arbiter uses the option-wise evidence audits and sufficiency reviews from all magnifications to produce the final answer and rationale:
\begin{equation}
    (\hat{y},\hat{r})=
    F_A\left(q,\left\{(m,\mathcal{A}_m,U_m)\right\}_{m\in\mathcal{M}}\right).
    \label{eq:final_evidence_adjudication}
\end{equation}
This evidence-level adjudication integrates morphological evidence across magnifications rather than using magnification-specific answers as votes. The answers $\{O_m\}_{m\in\mathcal{M}}$ are retained to preserve independent decisions across magnifications and trace the final decision.

\begin{table}[t]
  \caption{Zero-shot VQA accuracy (\%) for AdaptivePath and existing methods on SlideBench-VQA (BCNB), WSI-VQA, and the yes-or-no subset of PathVQA. The best and second-best results are shown in \textbf{bold} and \underline{underlined}, respectively.}
  \label{tab:zero-shot main}
  \centering
  \resizebox{\textwidth}{!}{%
  \begin{tabular}{l|cccccc|c|c}
    \toprule
    \multirow[b]{2}{*}{Method (Input)}
    & \multicolumn{6}{c|}{SlideBench-VQA (BCNB)}
    & WSI-VQA
    & PathVQA \\ 
    \cmidrule(lr){2-7}
    \cmidrule(lr){8-8}
    \cmidrule(lr){9-9}
    & \makecell{Overall\\Accuracy}
    & \makecell{Tumor\\Type}
    & \makecell{Receptor\\Status}
    & \makecell{HER2\\Expression}
    & \makecell{Histological\\Grading}
    & \makecell{Molecular\\Subtype}
    & Accuracy
    & Accuracy \\ 
    \midrule
    \rowcolor{gray!10}\multicolumn{9}{l}{\textbf{General-domain VLMs}} \\
    Qwen3-VL (Thumbnail)
    & 30.70 & 41.48 & 50.63 & 19.37 & 25.95 & 11.03 & 21.84 & - \\
    Qwen3-VL (Patch)
    & 38.22 & 43.96 & 53.26 & 21.35 & 29.74 & 16.04 & 23.47 & 44.82 \\
    GPT-5.5 (Thumbnail)
    & 37.36 & 40.72 & 52.62 & 20.15 & 26.73 & 19.35 & 25.48 & - \\
    GPT-5.5 (Patch)
    & 40.32 & 45.58 & 53.05 & 22.82 & 30.68 & 20.16 & 28.76 & 47.14 \\
    \midrule
    \rowcolor{gray!10}\multicolumn{9}{l}{\textbf{Medical-domain VLMs}} \\
    LLaVA-Med (Thumbnail)
    & 0.01 & 0.01 & 0.01 & 0.00 & 0.00 & 0.00 & 13.20 & - \\
    LLaVA-Med (Patch)
    & 30.10 & 23.95 & 42.52 & 23.72 & 18.99 & 15.05 & 21.55 & 27.78 \\
    MedDr (Thumbnail)
    & 35.48 & 28.92 & 48.08 & 20.65 & 29.96 & 23.88 & 43.69 & - \\
    MedDr (Patch)
    & 33.67 & 45.46 & 40.44 & 22.73 & 30.28 & 15.49 & 45.42 & 29.35 \\
    Quilt-LLaVA (Thumbnail)
    & 41.55 & 67.41 & 55.33 & 15.97 & 22.89 & 16.27 & 26.17 & - \\
    Quilt-LLaVA (Patch)
    & 44.43 & 77.14 & 56.46 & 23.18 & 18.23 & 19.82 & 29.23 & 20.76 \\
    \midrule
    \rowcolor{gray!10}\multicolumn{9}{l}{\textbf{Patch-aggregated Methods}} \\
    WSI-VQA (Slide)
    & 23.35 & 3.90 & 40.82 & 10.53 & 30.00 & 0.00 & 46.90 & 33.59 \\
    SlideChat (Slide)
    & 54.14 & \textbf{90.17} & \underline{73.09} & 25.05 & 23.11 & 17.49 & - & 55.03 \\
    WSI-LLaVA (Slide)
    & 52.68 & 85.32 & 60.89 & 24.75 & 46.28 & 29.20 & 45.71 & 54.97 \\
    \midrule 
    \rowcolor{gray!10}\multicolumn{9}{l}{\textbf{Agentic Systems}} \\
    GIANT (Slide)
    & 44.97 & 57.37 & 42.55 & 14.79 & 48.52 & 26.63 & 46.27 & 55.01 \\
    PathAgent (Slide)
    & 55.72 & 87.52 & 61.33 & 25.34 & \underline{55.95} & 30.21 & 51.54 & 58.36 \\
    PathNavigate (Slide)
    & \underline{59.42} & 88.28 & 65.97 & \underline{26.47} & 52.58 & \underline{32.33} & \underline{52.21} & \underline{58.94} \\
    \textbf{AdaptivePath (Slide)}
    & \textbf{60.81} & \underline{89.58} & \textbf{75.22} & \textbf{27.92} & \textbf{56.95} & \textbf{32.68} & \textbf{54.92} & \textbf{60.73} \\

    \bottomrule
  \end{tabular}
  }
\end{table}

\section{Experiments}

\subsection{Abnormal-Region Recognition Dataset}
\label{sec:dataset-construction}
To train the Navigator's active perception policy, we construct an abnormal-region recognition dataset comprising 6,352 pathology image patches from 880 WSIs across 30 TCGA cancer types. Patho-R1-7B~\cite{pathor1} generates the binary labels for normal and abnormal regions. Six pathologists, each with five years of pathology experience, review and correct these labels. We split the dataset into training, validation, and test sets in an 8:1:1 ratio, with no WSI shared across the splits. The dataset does not overlap with any downstream evaluation dataset. Multi-magnification sampling exposes the Navigator to tissue architecture and cellular morphology at different resolutions, facilitating its learning of observation locations and spatial extents.

\begin{table}[t]
  \caption{Zero-shot cancer subtype classification performance of AdaptivePath and existing methods on six TCGA cohorts. Results are reported as accuracy (\%). The best performance is shown in \textbf{bold}, while the second-best performance is \underline{underlined}.}
  \label{tab:tcga_zero_shot_classification}
  \centering
  \resizebox{\textwidth}{!}{%
\begin{tabular}{l|cccccc|c}
    \toprule
    Method (Input)
    & TCGA-BRCA
    & TCGA-BLCA
    & TCGA-ESCA
    & TCGA-NSCLC
    & TCGA-THCA
    & TCGA-RCC
    & Overall\\
    \midrule
    \rowcolor{gray!10}\multicolumn{8}{l}{\textbf{General-domain VLMs}} \\
    Qwen3-VL (Patch)
    & 32.84 & 38.71 & 37.50 & 56.47 & 34.13 & 15.96 & 41.00 \\
    GPT-5.5 (Patch)
    & 35.82 & 43.55 & 43.75 & 55.04 & 26.95 & 21.28 & 40.43 \\
    \midrule
    \rowcolor{gray!10}\multicolumn{8}{l}{\textbf{Medical-domain VLMs}} \\
    LLaVA-Med (Patch)
    & 49.25 & 56.45 & 46.88 & 37.05 & 40.72 & 48.94 & 42.86 \\
    Quilt-LLaVA (Patch)
    & 44.78 & 37.10 & 34.38 & 48.56 & 47.31 & \underline{56.38} & 47.29 \\
    \midrule
    \rowcolor{gray!10}\multicolumn{8}{l}{\textbf{Patch-aggregated Methods}} \\
    WSI-VQA (Slide)
    & 40.30 & 46.77 & 53.13 & 43.53 & 47.90 & 22.34 & 42.14 \\
    SlideChat (Slide)
    & 46.27 & \underline{62.90} & \underline{84.38}
    & 70.86 & 71.26 & 46.81 & \underline{65.29} \\
    WSI-LLaVA (Slide)
    & 52.24 & 56.45 & 68.75
    & \underline{73.02} & 68.26 & 30.85 & 62.57 \\
    \midrule
    \rowcolor{gray!10}\multicolumn{8}{l}{\textbf{Agentic Systems}} \\
    GIANT (Slide)
    & 31.34 & 46.77 & 65.63 & 48.92 & 61.08 & 27.66 & 47.86 \\

    PathAgent (Slide)
    & 47.76 & 53.23 & 75.00 & 69.06
    & \underline{71.86} & 35.11 & 62.00 \\

    PathNavigate (Slide)
    & \underline{53.73} & 51.61 & 81.25
    & 69.78 & 67.07 & 38.30 & 62.29 \\

    \textbf{AdaptivePath (Slide)}
    & \textbf{70.15}
    & \textbf{66.13}
    & \textbf{96.88}
    & \textbf{84.53}
    & \textbf{82.63}
    & \textbf{73.40}
    & \textbf{80.14} \\

    \bottomrule
  \end{tabular}
  }
\end{table}

\subsection{Implementation Details}
We implement AdaptivePath in PyTorch 2.7.0 and run experiments on NVIDIA GeForce RTX 3090 GPUs. The Navigator is optimized with AdamW~\cite{adam,adamw} using a batch size of 32, a weight decay of 0.02, and gradient clipping at 0.5. For the representation and PPO objectives in Eqs.~\ref{eq:repr_loss} and~\ref{eq:ppo_loss}, we set $\alpha=1$, the PPO clipping threshold to $0.2$, and $\beta_{\mathrm{stab}}=\beta_{\mathrm{ent}}=0.01$. Phase I runs for 300 epochs with cosine learning rate decay from $1\times10^{-4}$ to $1\times10^{-6}$. The Phase I checkpoint with the highest validation accuracy initializes Phase II, which runs for 100 epochs with an initial learning rate of $5\times10^{-5}$ and other settings unchanged. We set the region input size to $P=512$ and the glance and focus input sizes to $P_f=112$.

During inference, the Navigator uses magnifications $\mathcal{M}=\{2.5\times,5\times,10\times\}$, and each candidate region is resized to $P\times P$. We set the region budget to $K=5$ at each magnification and the focus budget to $T=4$ for each selected region. $K$ and $T$ define the observation budget for active perception. Unless otherwise specified, Patho-R1-7B~\cite{pathor1} serves as the Interpreter, while Qwen3-8B~\cite{qwen3} is the Deliberator and Arbiter. Both models remain frozen.

\subsection{Evaluation Datasets and Metrics}
We evaluate AdaptivePath on WSI-level and region-level pathology VQA and zero-shot cancer subtype classification of WSIs. SlideBench-VQA~\cite{slidechat} and WSI-VQA~\cite{wsi-vqa} require question answering over complete WSIs and therefore evaluate the full AdaptivePath workflow from active perception to evidence adjudication. PathMMU~\cite{pathmmu} and PathVQA~\cite{pathvqa} provide questions associated with pathology regions. For PathVQA, we report the subset whose answers are yes or no, while the PathMMU results are provided in the Supplementary Materials. The zero-shot cancer subtype classification evaluation comprises 700 WSIs from six TCGA cohorts. AdaptivePath uses only hematoxylin and eosin (H\&E)-stained WSIs for receptor status prediction and cancer subtype classification, with no immunohistochemistry (IHC) images or molecular assay results provided as input. We report percentage accuracy for all tasks. Dataset statistics and evaluation protocols are provided in the Supplementary Materials.

\subsection{Performance Comparisons}
To evaluate AdaptivePath's active perception workflow, we compare it with four baseline groups: general-domain VLMs, including Qwen3-VL~\cite{qwen3vl} and GPT-5.5~\cite{gpt5.5}; medical-domain VLMs, including LLaVA-Med~\cite{llavamed}, MedDr~\cite{meddr} and Quilt-LLaVA~\cite{quiltllava}; patch-aggregated methods, including WSI-VQA~\cite{wsi-vqa}, SlideChat~\cite{slidechat} and WSI-LLaVA~\cite{wsi-llava}; and agentic systems including GIANT~\cite{giant}, PathAgent~\cite{pathagent} and PathNavigate~\cite{pathnavigate}.

Table~\ref{tab:zero-shot main} reports the zero-shot VQA results. Since VLMs have constraints on image scale, we evaluate them using the following strategies~\cite{slidechat}: (1) ``Thumbnail'' resizes the complete WSI to $1024\times1024$. (2) ``Patch'' randomly samples 30 patches at $20\times$ and combines their predictions by majority voting. For PathVQA, we directly use the original pathology image for each question.

AdaptivePath achieves accuracies of 54.92\% on WSI-VQA and 60.81\% on SlideBench-VQA (BCNB), surpassing PathNavigate by 2.71 and 1.39 percentage points, respectively. It also achieves the highest PathVQA accuracy of 60.73\%. The gains on both WSI benchmarks indicate the effectiveness of combining active perception with evidence deliberation and final adjudication for whole-slide VQA.

Table~\ref{tab:tcga_zero_shot_classification} reports zero-shot cancer subtype classification results across six TCGA cohorts. AdaptivePath achieves accuracy of 80.14\%, exceeding the strongest baseline, SlideChat, by 14.85 percentage points. It ranks first on six cohorts, with margins over the strongest baseline ranging from 3.23 to 17.02 percentage points. These results demonstrate that the complete AdaptivePath workflow, from active perception to evidence adjudication, also transfers to cancer classification without cancer subtype annotations for training.

\subsection{Ablation Study}
Table~\ref{tab:vlm_llm_ablation} compares different Interpreter and Deliberator/Arbiter\cite{deepseekr1,llama3} combinations. Patho-R1-7B achieves the strongest results when paired with the Qwen3 models, and Qwen3-14B provides the highest accuracy on both datasets. Qwen3-8B closely matches this performance while reducing inference time from 212.56 to 122.26 seconds per question. We therefore use Patho-R1-7B and Qwen3-8B by default.

\begin{figure}[htbp]
    \centering
    \includegraphics[width=1\linewidth]{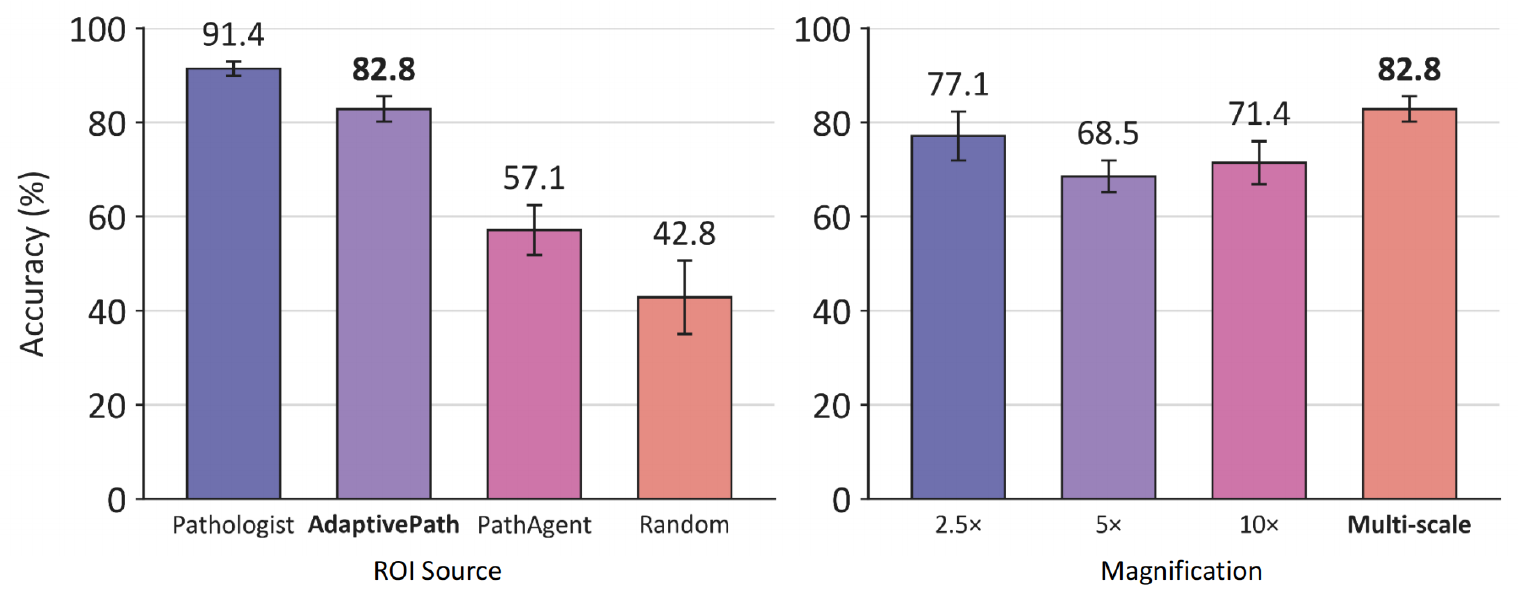}
    \caption{Diagnostic utility test on a subset of SlideBench-VQA (BCNB) under the same ROI budget. The left panel compares navigation sources, while the right compares individual magnifications with multi-scale evidence.}
    \label{fig:Diagnostic_utility_test}
\end{figure}

\begin{figure}[htbp]
    \centering
    \includegraphics[width=1\linewidth]{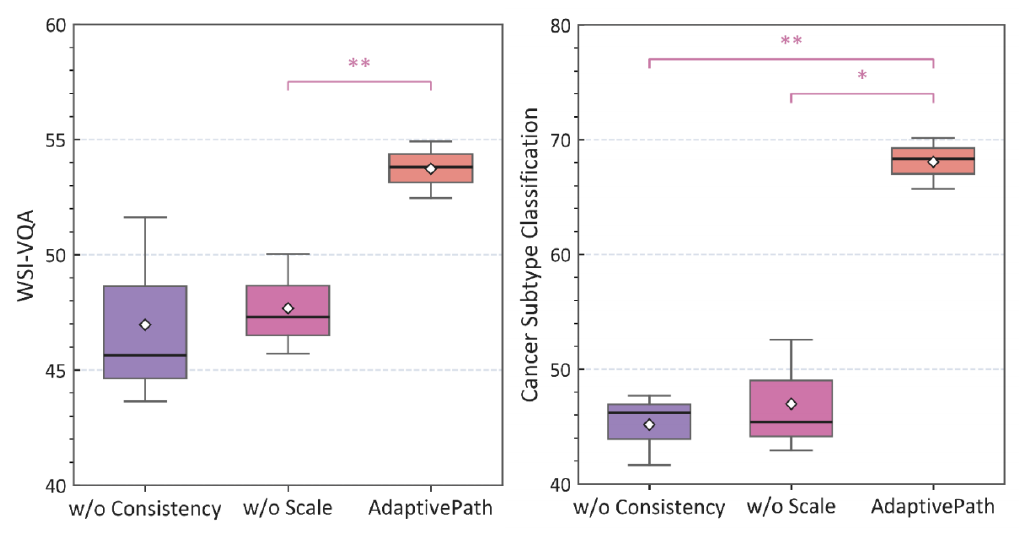}
    \caption{Navigator ablations on WSI-VQA and cancer classification. Asterisks indicate significance under paired $t$-tests ($^{*}p<0.05$, $^{**}p<0.01$).}
    \label{fig:Ablation_fig}
\end{figure}

Figure~\ref{fig:Ablation_fig} shows ablations of Phase II consistency regularization and the learnable scale action in the Navigator. Removing Phase II consistency regularization or replacing the learnable scale action with a fixed scale reduces performance on both WSI-VQA and cancer classification. AdaptivePath significantly outperforms the variant without consistency fine-tuning ($p<0.01$) and without scale action ($p<0.05$). These results validate consistency-regularized fine-tuning and scale-adaptive focus selection.

\begin{table}[t]
\caption{Ablations of ROI selection, magnification, and ROI budget under two Interpreter and Deliberator/Arbiter configurations. \textbf{Bold} indicates the best result in each column.}
\label{tab:module_ablation}
\centering
\resizebox{1\linewidth}{!}{
\begin{tabular}{l|cc|cc}
\toprule
\multirow{2}{*}{Variant} 
& \multicolumn{2}{c|}{Patho-R1-7B + Qwen3-8B} 
& \multicolumn{2}{c}{Quilt-LLaVA + Llama-3.1-8B}
\\
\cmidrule(lr){2-5}
& SlideBench-VQA (BCNB)
& WSI-VQA
& SlideBench-VQA (BCNB)
& WSI-VQA  
\\
\midrule
\rowcolor{gray!10}
\multicolumn{5}{l}{\textit{A. ROI selection}} \\
Random selection & 40.53 & 39.85 & 40.33 & 39.56 \\
PLIP selection   & 44.38 & 45.30 & 43.92 & 45.72 \\
CONCH selection  & 48.95 & 46.76 & 47.16 & 45.94 \\

\midrule
\rowcolor{gray!10}
\multicolumn{5}{l}{\textit{B. Magnification used}} \\
2.5$\times$ only & 52.73 & 52.47 & 52.88 & 51.65 \\
5$\times$ only   & 55.81 & 50.46 & 55.29 & 50.26 \\
10$\times$ only  & 51.64 & 52.75 & 51.48 & 52.05 \\

\midrule
\rowcolor{gray!10}
\multicolumn{5}{l}{\textit{C. ROI budget}} \\
Top-1 & 58.77 & 51.26 & 55.89 & 50.59 \\
Top-3 & 59.03 & 52.71 & 57.61 & 51.76 \\
\rowcolor{gray!10}
\textbf{Top-5 (Ours)} & \textbf{60.81} & \textbf{54.92} & \textbf{59.58} & \textbf{52.93} \\
Top-8  & 59.27 & 52.53 & 59.11 & 52.34 \\
Top-10 & 58.84 & 50.26 & 57.65 & 50.28 \\
\bottomrule
\end{tabular}}
\end{table}

The navigation ablations in Table~\ref{tab:module_ablation} examine aspects of active perception: how ROIs are selected, which magnifications are used, and how many ROIs are retained. Random sampling and retrieval with PLIP or CONCH reduce accuracy in both module configurations relative to the trained Navigator, supporting the benefit of abnormality-supervised navigation. The strongest magnification varies between the two datasets, whereas combining all magnifications yields the highest accuracy in every column. Region budget of $K=5$ performs best across both module configurations, and larger budgets provide no improvement. We therefore use all magnifications and $K=5$ as the default navigation setting.

\begin{table}[t]
\caption{Comparison of accuracy (\%) and inference time across AdaptivePath configurations. Inference time is reported in seconds per question.}
\label{tab:vlm_llm_ablation}
\centering
\resizebox{1\linewidth}{!}{
\begin{tabular}{ll|cc|c}
\toprule
Interpreter
& Deliberator / Arbiter
& SlideBench-VQA (BCNB)
& WSI-VQA
& Time(s/question) \\
\midrule

\multirow{4}{*}{Patho-R1-7B}
& Qwen3-8B       & 60.81 & 54.92 & 122.26 \\
& Qwen3-14B      & 60.98 & 55.03 & 212.56 \\
& Llama-3.1-8B   & 58.34 & 53.06 & 131.68 \\
& DeepSeek-R1-8B & 57.53 & 52.47 & 161.64 \\

\midrule

\multirow{4}{*}{Quilt-LLaVA}
& Qwen3-8B       & 59.74 & 53.18 & 119.43 \\
& Qwen3-14B      & 60.17 & 54.62 & 225.58 \\
& Llama-3.1-8B   & 59.58 & 52.93 & 128.95 \\
& DeepSeek-R1-8B & 58.22 & 53.25 & 152.46 \\

\bottomrule
\end{tabular}}
\end{table}

\subsection{Evaluation with Pathologists}
We assess whether acquired image sequences support diagnosis and whether navigation reaches regions examined during pathologist review. The Diagnostic Utility Test and Trajectory Alignment Experiment address these two questions, respectively. Detailed protocols and reader characteristics are provided in the Supplementary Materials.

\subsubsection{Diagnostic Utility Test}
In this blinded test, pathologists receive an anonymized raw-image sequence from one of four sources: a recorded pathologist trajectory, AdaptivePath, PathAgent, or matched random tissue sampling. All conditions use the same ROI budget, and the pathologists answer solely from the provided images. The reader study therefore measures the diagnostic utility of H\&E image sequences without ancillary testing. We compare single-magnification observations with the complete multi-scale sequence to assess the contribution of multi-scale evidence.

The left panel of Figure~\ref{fig:Diagnostic_utility_test} shows that sequences produced by AdaptivePath yield a mean accuracy of 82.9\%, compared with 50.9\% for PathAgent and 34.3\% for random sampling. Among automated methods, this is closest to the 89.7\% mean accuracy achieved with pathologist trajectories, supporting the diagnostic utility of actively acquired observations.

\begin{figure}[htbp]
    \centering
    \includegraphics[width=1\linewidth]{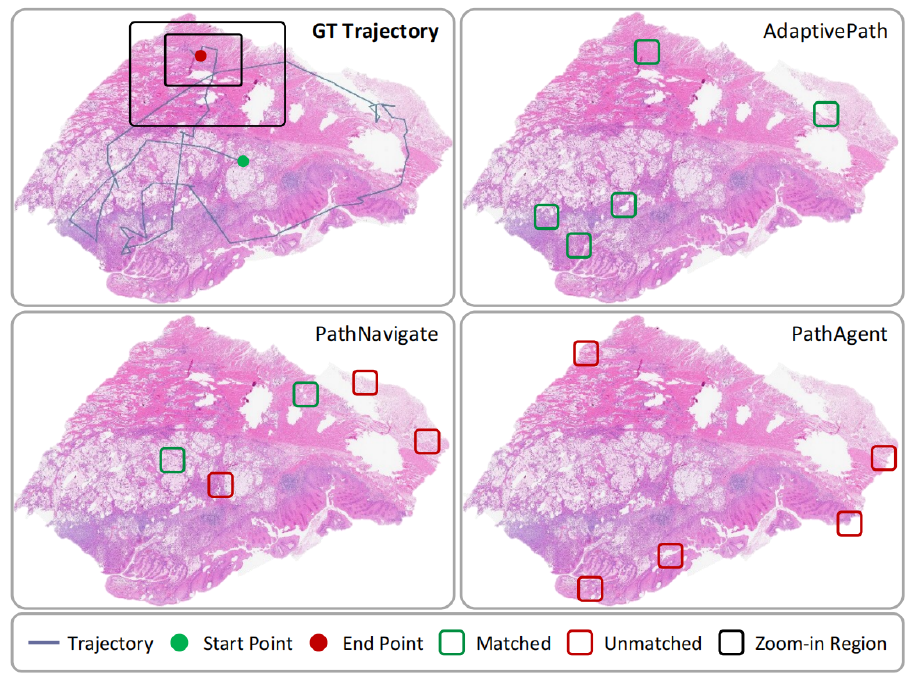}
    \caption{Qualitative navigation comparison under the same ROI budget. The blue line and black boxes show the pathologist's viewport-center trajectory and representative viewports, with green and red circles marking its start and end. Green and red boxes indicate selected ROIs with and without overlap with pathologist-inspected regions, respectively.}
    \label{fig:navigation_comparison}
\end{figure}

As shown in the right panel of Figure~\ref{fig:Diagnostic_utility_test}, the complete multi-scale sequence achieves a mean accuracy of 82.9\%, compared with 69.7\%, 62.9\%, and 62.3\% when evidence is restricted to $2.5\times$, $5\times$, and $10\times$, respectively. None of the individual magnifications matches the diagnostic utility of the complete sequence. This reader-study result suggests that a single magnification may omit complementary diagnostic evidence and supports integrating tissue architecture, regional organization, and cytological detail across magnifications.

\subsubsection{Trajectory Alignment Experiment}
The dataset contains 58 WSIs across five organs, each WSI annotated with a pathologist's navigation trajectory recorded during diagnostic review. All methods are evaluated under the same ROI budget. Figure~\ref{fig:navigation_comparison} provides a qualitative comparison. In the illustrated case, all AdaptivePath selections overlap pathologist-inspected regions, whereas the other methods exhibit partial or no overlap. Quantitative results across 58 WSIs are provided in the Supplementary Materials.

\section{Conclusion}
We presented AdaptivePath, an active-perception framework for agentic visual reasoning over gigapixel whole-slide images. By coupling scale-progressive evidence acquisition with explicit adjudication, AdaptivePath provides a traceable basis for whole-slide reasoning. To ground this process in pathology, we constructed a pathologist-reviewed abnormal-region recognition dataset and trained a scale-adaptive Navigator to determine where and over what spatial extent to observe. Experiments on pathology VQA and zero-shot cancer subtype classification demonstrate its effectiveness across tasks, while diagnostic-utility and trajectory-alignment evaluations further show that its observation sequences provide meaningful and traceable evidence. Since AdaptivePath currently operates only on H\&E-stained WSIs, findings requiring IHC or molecular testing remain outside its directly observable evidence. Future work will incorporate paired IHC slides and molecular data, investigate question-aware reranking, improve navigation efficiency, and extend AdaptivePath to more diverse pathological findings.

\bibliographystyle{plainnat}
\bibliography{references}


\end{document}